\documentclass[11pt]{article}
\usepackage{acl2015}
\usepackage{times}
\usepackage{url}
\usepackage{latexsym}
\usepackage{helvet}
\usepackage{courier}
\usepackage{graphicx}
\usepackage{multirow}
\usepackage{multicol}
\usepackage{amsmath}
\usepackage{epstopdf}
\usepackage{amsfonts,amssymb}
\usepackage{float}
\usepackage{hhline}
\usepackage{subfig}
\usepackage{footnote}
\usepackage{array}
\usepackage{bm}
\makesavenoteenv{tabular}
\makesavenoteenv{table}

\title{A Dependency-Based Neural Network for Relation Classification}

\author{
Yang Liu$^{1}$\thanks{~~Contribution during internship at Microsoft Research.} ~~ Furu Wei$^2$ ~~ Sujian Li$^{1}$ ~~ Heng Ji$^3$ ~~ Ming Zhou$^2$~~ Houfeng Wang$^{1}$\\
  $^1$Key Laboratory of Computational Linguistics, Peking University, MOE, China \\
  $^2$Microsoft Research, Beijing, China\\
  $^3$Computer Science Department, Rensselaer Polytechnic Institute, Troy, NY, USA\\
  {\tt \{cs-ly, lisujian, wanghf\}@pku.edu.cn } \\
  {\tt \{furu, mingzhou\}@microsoft.com }
  {\tt jih@rpi.edu }
  }
\date{}

\begin{document}
\maketitle
\begin{abstract}
Previous research on relation classification has verified the effectiveness of using dependency shortest paths or subtrees.
In this paper, we further explore how to make full use of the combination of these dependency information.
We first propose a new structure, termed augmented dependency path (ADP), which is composed of the shortest dependency path  between two entities  and the subtrees attached to the shortest path.
To exploit the semantic representation behind the ADP structure, we develop   dependency-based neural networks (DepNN): a recursive neural network  designed to model the subtrees, and a convolutional neural network to capture the most important features on the shortest path.
Experiments on the SemEval-2010 dataset show that our proposed method achieves  state-of-art results.

\end{abstract}

\section{Introduction}
Relation classification aims to classify the semantic relations between two entities in a sentence. It plays a vital role in robust knowledge extraction from unstructured texts and serves as an intermediate step in a variety of natural language processing
applications. Most existing approaches  follow a machine learning based framework and focus on
designing effective features to obtain better classification performance.

The effectiveness of using dependency relations between entities for relation classification has been reported in previous approaches~\cite{bach2007survey}.
For example, ~\newcite{suchanek2006combining} carefully selected a set of features from tokenization and dependency parsing, and extended some of them to generate high order features in different ways.
\newcite{1766597} designed a dependency tree kernel and attached more information including Part-of-Speech tag, word chunking tag to each node in the tree. Interestingly, \newcite{2363908} provided an important insight that the shortest path between  two entities in a dependency graph concentrates most of the information for identifying the relation between them.
\newcite{nguyen2007relation} developed these ideas by analyzing multiple subtrees with the guidance of pre-extracted keywords.
 Previous work showed that the most useful dependency information in relation classification includes the shortest dependency path and dependency subtrees. These two kinds of information serve different functions and their collaboration can boost the performance of relation classification  (see Section 2 for detailed examples).
However, how to uniformly and efficiently  combine these two components is still an open problem.
In this paper, we propose a novel structure named Augmented Dependency Path (ADP) which attaches dependency subtrees to words on a shortest dependency path and focus on exploring the semantic representation behind the ADP structure.

\begin{figure*}
\centering
\centering\includegraphics[width=4.5in]{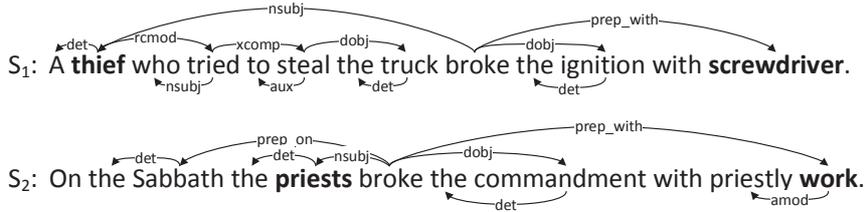}
\caption{Sentences and their dependency trees.}
\label{fig1}
\end{figure*}

\begin{figure*}
\centering
\subfloat[Augmented dependency path for $S_1$.]{
\label{fig:2_subfig_a}
\centering
\includegraphics[height=1.1in]{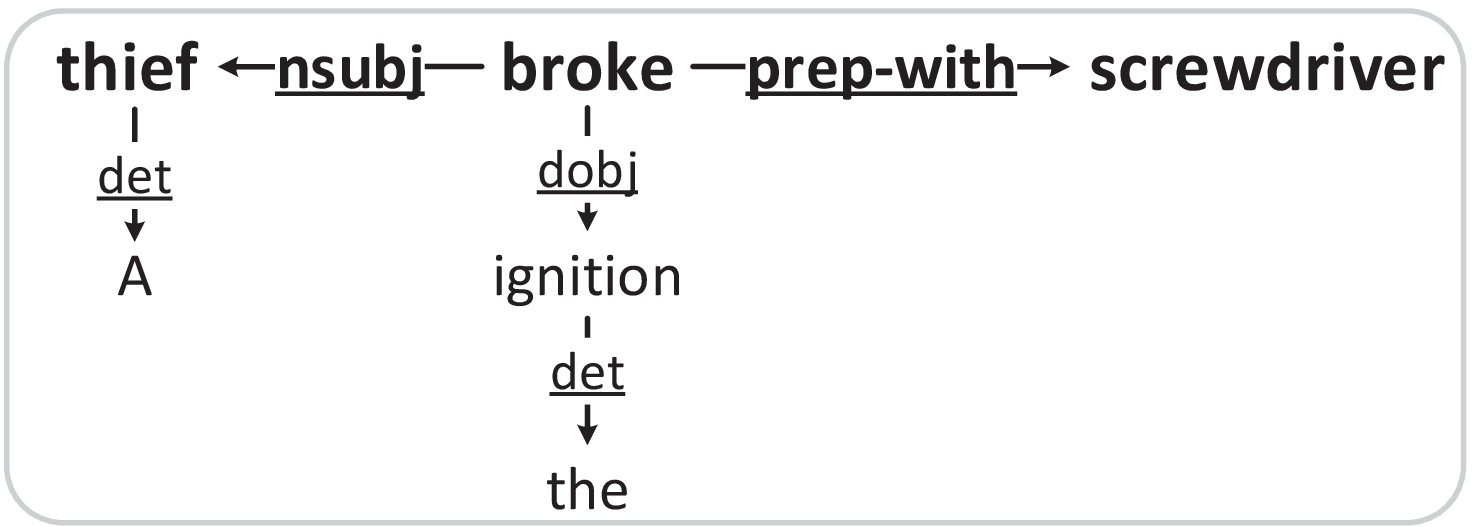}
}
\centering
\subfloat[Augmented dependency path for $S_2$.]{
\label{fig:2_subfig_b}
\centering
\includegraphics[height=1.1in]{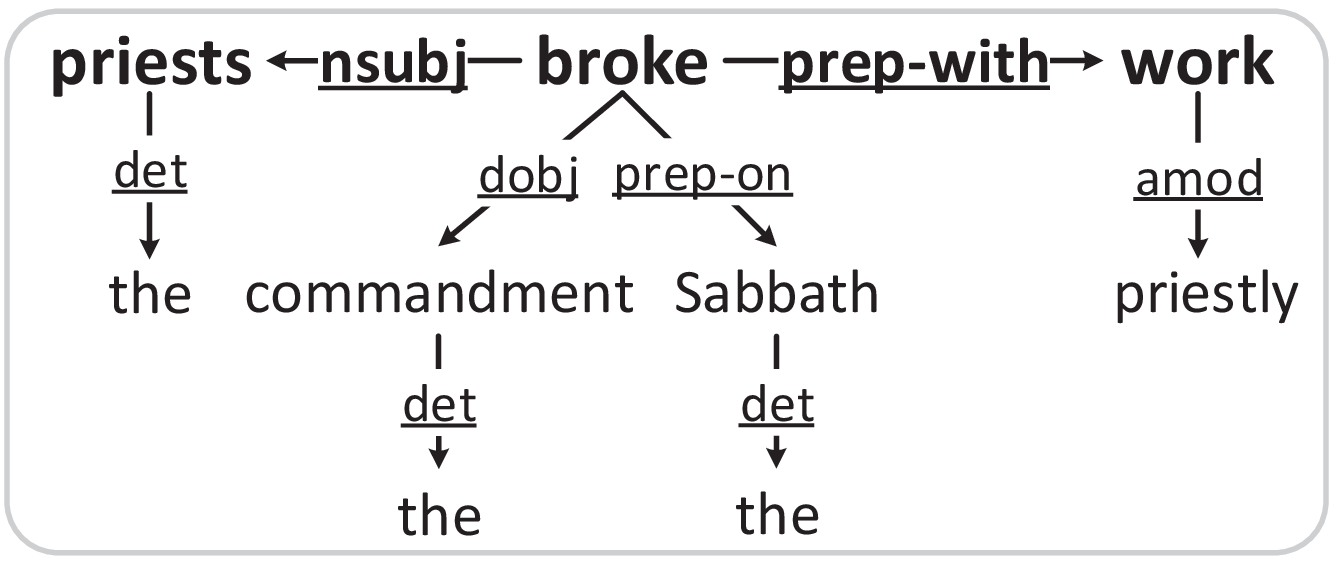}
}
\caption{ The bold part is the shortest path between two entities in the undirected version of dependency tree, and some subtrees are attached to it. They two are combined as an augmented dependency path. }
\label{fig1}
\end{figure*}

Recently, deep learning techniques have been widely used in modeling complex structures. This provides us an opportunity to model the ADP structure in a neural network framework.
Thus, we propose a  dependency-based neural network where two sub-neural networks are used to model shortest dependency paths and dependency subtrees respectively.
One convolutional neural network (CNN) is applied over the shortest dependency path, because CNN is suitable  for capturing the most useful features in a flat structure.
A recursive neural network (RNN) is used for extracting semantic representations from the dependency subtrees, since RNN  is good at modeling hierarchical structures.
To connect these two sub-networks, each word on the shortest path is combined with a representation generated from its subtree, strengthening the semantic representation of the shortest path.
In this way, the augmented dependency path is represented as a continuous semantic vector which can be further used for relation classification.

The major contributions of the work presented in this paper are as follows.
\begin{enumerate}
\item We extend the shortest dependency path into the augmented dependency path to better model the relation between two entities.
\item We propose a dependency-based neural network, DepNN, to model the augmented dependency path. It combines the advantages of the convolutional neural network and the recursive neural network.
\item We conduct extensive experiments on the SemEval 2010 dataset and the experimental results show that DepNN outperforms baseline methods and yields state-of-the-art F1 measure on the relation classification task.
\end{enumerate}

\section{Problem Definition and Motivation}

\begin{figure*}[htbp]

\centering\fbox{\includegraphics[width=5.5in,trim=2 2 2 2,clip]{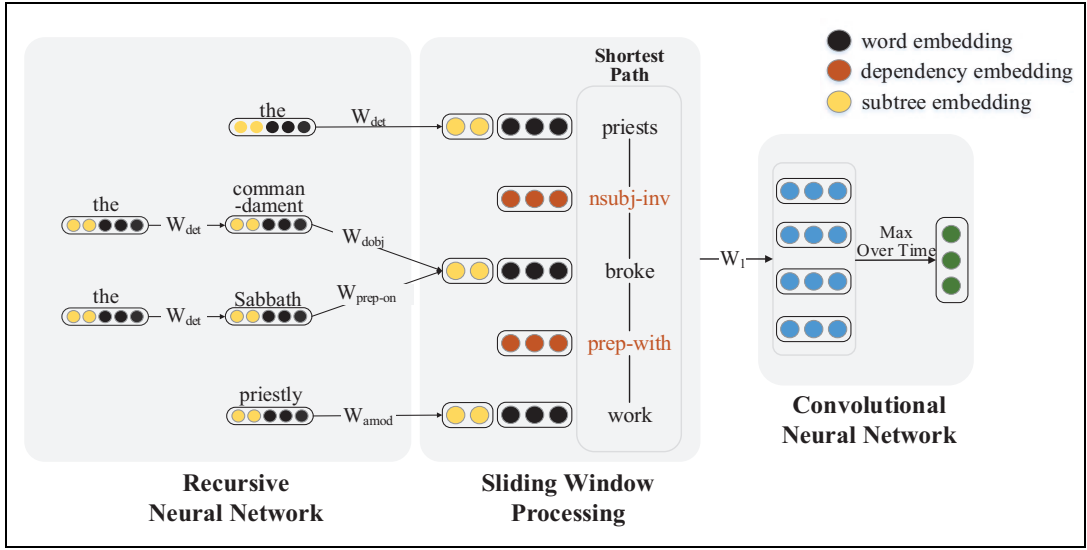}}
\caption{ Illustration of dependency-based neural networks. }\label{fig:4}
\end{figure*}

The task of relation classification can be defined as follows. Given a sentence $S$ with a pair of entities $e_1$ and $e_2$ annotated, the task is to identify the semantic relation between $e_1$ and $e_2$ in accordance with a set of predefined relation types. According to the the official guideline of SemEval-2010 task 8~\cite{5009814}, there are 9 ordered relation types.  We list them in Table \ref{tab:types} with their simplified definitions. Instances don't fall in any of these types are labeled as \textit{Other}. For example, in Figure \ref{fig1}, the relation between two entities $e_1$=\emph{thief} and $e_2$=\emph{screwdriver} is \textit{Instrument-Agency}.

\begin{table}
\renewcommand\arraystretch{1.1}
\small
\begin{tabular}{|l|p{15em}|}
\hline
Relation Type & Definition\\
\hline
Cause-Effect&$X$ is the cause of $Y$\\
\hline
Entity-Origin&$Y$ is the origin of an entity $X$ , and $X$ is coming
or derived from that origin. \\
\hline
Message-Topic&$X$ is a communicative message containing information about $Y$\\
\hline
Product-Producer&$X$ is a product of $Y$\\
\hline
Entity-Destination&$Y$ is the destination of $X$ in the sense of $X$ moving toward $Y$\\
\hline
Member-Collection&$X$ is a member of $Y$\\
\hline
Instrument-Agency&$X$ is the instrument (tool) of $Y$ or $Y$ uses $X$\\
\hline
Component-Whole&$X$ has an operating or usable purpose within $Y$\\
\hline
Content-Container&$X$ is or was stored or carried inside $Y$ \\
\hline
\end{tabular}
\caption{\label{tab:types}Relation types of $(X,Y)$ and their definitions in SemEval-2010 task 8.}
\end{table}

\newcite{2363908} reported that, for the relation classification task,  the shortest dependency path between two entities plays a vital role. They pointed out that this kind of paths can capture the predicate-argument sequences, providing helpful information for relation classification. For example, in Figure \ref{fig:2_subfig_a}, the shortest path includes the structure of ``broke \textit{prep\_with} screwdriver'', helping judging the \textit{Instrument-Agency} relation.

Although the shortest dependency paths prove useful for relation classification, there exists other information on the dependency tree that can be exploited to represent the relation  more precisely.
For example, Figure \ref{fig:2_subfig_a} and \ref{fig:2_subfig_b} show two instances which have similar shortest dependency paths but belong to different relation types.
In this situation, if we only use the shortest dependency paths for judging relation types, it is difficult for us to distinguish these two instances.
However, we notice that the  subtrees attached to the shortest dependency paths such as  ``dobj$\rightarrow$commandment'' and ``dobj$\rightarrow$ignition'' can provide supplemental information for relation classification.
Based on many observations like this, we propose the idea that we should employ these subtrees and combine them
with the shortest  path to form a more precise structure for classifying  relations.
 This combined structure is called ``\textbf{augmented dependency path} (ADP)'', as illustrated in Figure~\ref{fig1}.

Next, our goal is to capture the semantic representation of the ADP structure between two entities.
The key problem here is how to combine the two components of ADP to incorporate more information.  We propose that on the augmented dependency path, a word should be represented by both itself and its attached subtree.
This is because the word itself contains its general  meaning while the subtree can provide semantic information about how this word functions in this specific sentence.
With this idea, we adopt the recursive neural network (RNN) that is proved suitable for modeling hierarchical structures to build semantic embeddings for the words on the shortest path along with their subtrees.
After obtaining these more precise word representations, a convolutional neural network (CNN) can be applied, since it is good at modeling flat structures and can generate a fix-sized vector containing the most relevant features.

\section{Dependency-Based Neural Networks}

In this section, we will introduce how we use neural network techniques and dependency information to explore the semantic connection between two entities.
We name our architecture of modeling ADP structures as dependency-based neural networks (DepNN).
Figure \ref{fig:4} illustrates DepNN with a concrete example.
First, we associate each word $w$ and dependency relation $r$  with a vector representation $\bm{x}_w, \bm{x}_r \in \mathbb{R}^{dim}$.
For each word $w$ on the shortest dependency path, we develop an RNN from its leaf words up to the root to generate a subtree embedding $\bm{c}_w$ and concatenate $\bm{c}_w$ with $\bm{x}_w$ to serve as the final representation of $w$.

Next, a CNN is designed to model the shortest dependency path based on the representation of its words and relations.
Finally our framework can efficiently represent the semantic connection between two entities with consideration of more comprehensive dependency information.

\subsection{Modeling Dependency Subtree}

The goal of modeling dependency subtrees is to find an appropriate representation for the words on the shortest path.
As mentioned above, we assume that each word $w$ can be interpreted by itself and its children on the dependency subtree.
Then, for each word $w$ on the subtree, its word embedding $\bm{x}_w \in \mathbb{R}^{dim}$ and subtree representation  $\bm{c}_w \in \mathbb{R}^{dim_{c}}$ are concatenated to form its final representation $\bm{p}_w \in \mathbb{R}^{dim+dim_{c}}$.
For a word that does not have a subtree, we set its subtree representation as $\bm{c}_{LEAF}$.
The subtree representation of a word is derived through transforming the representations of its children words.
During the bottom-up construction of the subtree, each word is associated with a dependency relation such as \textit{dobj} as in Figure~\ref{fig:4}.

Take the ADP in Figure \ref{fig:4} for example, we first compute leaves' representations like $p_{the}$,
\begin{equation}
\bm{p}_{the}=[\bm{x}_{the}, \bm{c}_{LEAF}]
\end{equation}

Once all leaves are finished, we move to interior nodes with
already processed children. In the example, continuing from ``the'' to its parent, ``Sabbath'', we compute
\begin{align}
\bm{p}_{Sabbath}&=
[\bm{x}_{Sabbath}, \bm{c}_{Sabbath}]\\
\bm{c}_{Sabbath}&=f(\bm{W}_{det} \cdot \bm{p}_{the} +\bm{b})
\end{align}
where $f$ is a non-linear activation function such
as $tanh$, $\bm{W}_{det}$ is the transformation matrix associated with dependency relation $det$ and $\bm{b}$ is a bias term. We repeat this process until we reach the root on the shortest path, which in this case is ``broke'',
\begin{align}
\nonumber \bm{p}_{broke}&=
\nonumber [\bm{x}_{broke}, \bm{c}_{broke}]\\
\nonumber \bm{c}_{broke}&=f(\bm{W}_{prep-on} \cdot \bm{p}_{Sabbath} \\
\nonumber &+\bm{W}_{dobj} \cdot \bm{p}_{commandament})
\end{align}

The composition equation for any word $w$ with
children $Q(w)$ is,
\begin{gather}
\bm{c}_w=f(\sum_{q \in Children(w)}\bm{W}_{R_{(w,q)}}\cdot \bm{p}_q+\bm{b}) \\
\bm{p}_q=[\bm{x}_{q}, \bm{c}_{q}]
\end{gather}
where $R_{(w, q)}$ denotes the dependency relation between word $w$ and its child word $q$.
This process continues recursively from leaves up to the root words on the shortest path. Each of these words will have a vector representation after this stage ($\bm{p}_{priests}$, $\bm{p}_{broke}$ and $\bm{p}_{work}$ in this example).

\subsection{Modeling Shortest Dependency Path}
To classify the relation between two entities, we further explore the semantic representation behind their shortest dependency path, which can be seen as  sequence of words interspersed with dependency relations.
Take the shortest dependency path in last subsection for example. The sequence $S$ will be,
\begin{center}
\includegraphics[width=2.9in]{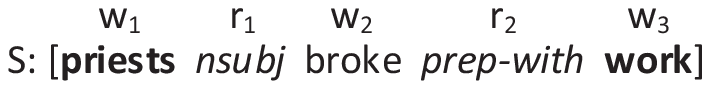}
\end{center}

As the convolutional neural network (CNN) is good at capturing the salient features from a sequence of objects, we design a CNN to tackle the shortest dependency path.

A CNN contains a convolution operation over  windows of object representations, followed by a pooling operation.
As we know, a word $w$ on the shortest path is associated with the representation $\bm{p}_w$ through modeling the subtree.
For a dependency relation $r$ on the shortest path, we set its representation as a vector $\bm{x}_r \in \mathbb{R}^{dim}$.
As a sliding window is applied on the sequence, we set the window size as $k$.
For example, when $k=3$, the sliding windows of $S$ are
$\{[r_s\ w_1\ r_1],[r_1\ w_2\ r_2], [r_2 \ w_3\ r_e]\}$
where $r_s$ and $r_e$ are used to denote the beginning and end of a shortest dependency path between two entities.

We concatenate $k$ neighboring word (or dependency relation) representations within one window into a new vector.
Assume $\bm{X}_i \in \mathbb{R}^{dim\cdot k + dim_c\cdot n_w}$ as the concatenated representation of the $i$-th window, where $n_w$ is the number of words in one window.
 A convolution operation involves a filter $\mathbf{W}_1 \in \mathbb{R}^{l \times (dim\cdot k+dim_c\cdot n_w)}$, which operates on $\bm{X}_i$  to produce a new feature vector $\bm{L}_i$ with $l$ dimensions,
 \begin{equation}
\bm{L}_i=\bm{W}_1\bm{X}_i
 \end{equation}
where the bias term is ignored for simplicity.

Then $\mathbf{W}_1$ is applied to each possible window in the shortest dependency path  to produce a feature map:
$[\bm{L}_0, \bm{L}_1, \bm{L}_2, \cdots ]$.
Next, we adopt the widely-used max-over-time pooling operation ~\cite{collobert2011natural}, which can retain the most important features, to obtain the final representation $L$  from the feature map.
That is, $\bm{L}=max(\bm{L}_0,\bm{L}_1,\bm{L}_2, \dots)$.

By this means, we are able to obtain the semantic representation of ADP with advantages of both RNN and CNN.

\subsection{Learning}
Like other relation classification systems, we also incorporate some lexical level features which are proved useful for this task. This includes named entity tags and WordNet hypernyms of $e_1$ and $e_2$. We concatenate them with the ADP representation $\bm{L}$ to produce a combined vector $\bm{M}$. We then pass $\bm{M}$ to a fully connected $softmax$ layer whose output is the probability distribution $\bm{y}$ over relation labels.
\begin{gather}
\bm{M}=[\bm{L}, \bm{LEX}] \\
\bm{y}=softmax(\bm{W}_2\bm{M})
\end{gather}
We define the ground-truth label  vector $\bm{t}$ for each instance as a binary vector. If the instance belongs to the the $i$-th type, only $\bm{t}_i$ is 1 and the other dimensions are set to 0. To learn the parameters, we optimize the cross-entropy error between $\bm{y}$ and $\bm{t}$ using stochastic gradient descent~\cite{bottou2004stochastic}. For each training instance, we define the objective function as:
\begin{gather}
\mathop{min}_{\theta}(-\sum_j^{l_n} \bm{t}_jlog(\bm{y}_j))
\end{gather}
where $\theta$ represents the parameters.
Gradients are computed using backpropagation~\cite{rumelhart1988learning}.

\section{Experiments}
Our experiments are performed on SemEval-2010 dataset~\cite{5009814}.
The training part of the dataset  includes 8000 instances, and the test part includes 2717 instances.
Table \ref{tab:stat} shows the statistics of the annotated relation types of this dataset.
We can see that the distribution of relation types in the test set is similar to that in the training set.
 The official evaluation metric is the macro-averaged F1-score (excluding \textit{Other}).
We use dependency trees generated by the Stanford Parser \cite{443415} with the ``collapsed'' option, which regards a preposition as a kind of dependency relation. As \newcite{4933584} pointed out, this option is more useful for event relation extraction.
\begin{table}[htbp]
\center
\small
\begin{tabular}{|c|cc|}
\hline
\multirow{2}{*}{Relation}&
		\multicolumn{2}{c|}{Frequency}\\
		\cline{2-3}
		& Train & Test\\
\hline
Other&1410 (17.63\%)&454 (16.71\%)\\
Cause-Effect&1003 (12.54\%)&328 (12.07\%)\\
Component-Whole& 941 (11.76\%)&312 (11.48\%)\\
Entity-Destination& 845 (10.56\%)&292 (10.75\%)\\
Product-Producer& 717 ( 8.96\%)&231 (8.50\%)\\
Entity-Origin& 716 ( 8.95\%)&258 (9.50\%)\\
Member-Collection& 690 ( 8.63\%)&233 (8.58\%)\\
Message-Topic& 634 ( 7.92\%)&261 (9.61\%)\\
Content-Container& 540 ( 6.75\%)&192 (7.07\%)\\
Instrument-Agency& 504 ( 6.30\%)&156 (5.74\%)\\
\hhline{|===|}
Total&8000 (100\%)&2717 (100.00\%)\\
\hline
 \end{tabular}
 \caption{Statistics of SemEval-2010 dataset.}\label{tab:stat}
\end{table}

\subsection{Analysis of DepNN}
\subsubsection{Contributions of different components}
We first show the contributions from different components of DepNN.
In our experiments, two kinds of word embeddings  are used for initialization. One is the 50-\textit{d} embeddings provided by SENNA~\cite{collobert2011natural}.
The second is the 200-\textit{d} embeddings~\cite{MoYu2014qv} trained on Gigaword with word2vec\footnote{https://code.google.com/p/word2vec/}.
The corresponding hyperparameters are set with 5-fold cross validation, including window size $k$, learning rate $\lambda$, subtree embedding's dimension $dim_c$, and hidden layer size $l$. The final settings are shown in Table \ref{tab:hyp}.
\begin{table}[htbp]
\center
\begin{tabular}{|c|p{0.8cm}<{\centering}|p{0.8cm}<{\centering}|p{0.8cm}<{\centering}|p{0.8cm}<{\centering}|}
\hline
&$k$&$\lambda$&$dim_c$&$l$\\
\hline
50-\textit{d}&5&0.05&25&200\\

200-\textit{d}&5&0.05&100&400\\
\hline
 \end{tabular}
 \caption{Hyperparameters settings.}\label{tab:hyp}
\end{table}

For evaluation, we first design a relation extraction system (named PATH) which only models the shortest dependency path with a CNN.
Based on PATH, We consider to incorporate the two kinds of lexical features including named entity tags (NER) and WordNet hypernyms (WN).
Then, we get two systems which are named PATH+WN and PATH+NER respectively.
We also add the attached subtrees (SUB) modeled by an RNN to form the complete augmented dependency path.
%Finally, all these components (PATH+SUB+WN+NER) are combined and DepNN achieves its best performance.

\begin{table}[!htbp]
	\center
	\begin{tabular}{|l|cc|}
		\hline
		\multirow{2}{*}{Model}&
		\multicolumn{2}{c|}{F1}\\
		\cline{2-3}
		& 50-\textit{d} & 200-\textit{d}\\
		\hline
		PATH&80.3&81.8\\
        PATH+WN&80.8&82.0\\
        PATH+NER&81.1&82.4\\
		PATH+SUB&81.2&82.8\\
		\hline
	\end{tabular}
	\caption{Performance of DepNN with different components.}\label{tab:feature sets}

\end{table}

From Table \ref{tab:feature sets}, we can verify the effectiveness of modeling the shortest dependency path with a CNN, since PATH can achieve a relatively high result. The experiment results also indicate that both the NER and WordNet features can improve the performance of relation extraction.
WordNet seems less useful than NER,  which conforms to the results of ~\newcite{MoYu2014qv} , since a large number of WordNet hypernyms may cause overfitting. Furthermore, the attached subtrees, as we expect, can provide an obvious boost to DepNN.
The NER tags, WordNet hypernyms and subtrees all contribute to the performance by providing supplemental information for words on the shortest path. The experiments show that the subtree information does a better job than the other two kinds of information and can help build more precise representations for words in a sentence.
To get a deeper understanding of what semantic information can be captured behind the ADP structure, we will look into our model and analyze it with specific examples. Since the Gigaword embeddings, with its larger corpus and dimensions, can significantly improve the classification performance, the following experiments and analysis are all based on Gigaword embeddings.

\subsubsection{Intuitive Analysis of Shortest Path}
We take the output vector of the CNN layer as the distributed representation of a dependency path. In this way, we can calculate the cosine similarity between any two paths and illustrate some paths with high similarity.
Table \ref{tab:4} shows three training instances with different relation types and their three most similar paths in the test set.

\begin{table}[h]
\small
\center
\renewcommand\arraystretch{1.2}
\begin{tabular}{|p{7.2cm}|}
\hhline{|=|}
 \multicolumn{1}{|c|}{Instrument-Agency}\\
 \hline
\textbf{master} \textit{nsubj\_inv} teaches \textit{dobj} lesson \textit{prep\_with} \textbf{stick} \\
\hline
\textbf{analyzer} \textit{prep\_of\_inv} core \textit{nsubj\_inv} identifies \textit{dobj} paths \textit{vmod} using \textit{dobj} \textbf{method} \\
\textbf{architect} \textit{nn\_inv} measures \textit{dep} Sage \textit{prep\_with} \textbf{strip}\\
\textbf{shop} \textit{nsubj\_inv} fixed \textit{prep\_with} \textbf{method}\\
\hhline{|=|}
\multicolumn{1}{|c|}{Product-Producer}\\
\hline
\textbf{factory} \textit{nsubj\_inv} began \textit{xcomp} manufacture  \textit{dobj} \textbf{banduras}\\
\hline
 \textbf{designer} \textit{nsubj\_inv} made \textit{dobj} \textbf{sets} \\
 \textbf{writer} rcmod pencilled \textit{dobj} \textbf{storyboard}\\
 \textbf{student} \textit{nsubj\_inv} spent \textit{xcomp} creating \textit{dobj} \textbf{application}\\
\hhline{|=|}
\multicolumn{1}{|c|}{Message-Topic}\\
\hline
\textbf{article} \textit{prep-in-inv} explores \textit{dobj} \textbf{impulsivity} \\
\hline
\textbf{article} \textit{rcmod} pointed \textit{dobj} \textbf{problems} \\
\textbf{speech} \textit{vmod} addressing \textit{dobj}\textbf{practices}\\
\textbf{chapter} \textit{nsubj\_inv} relates \textit{dobj} \textbf{attempts}\\
\hline
\end{tabular}
\caption{Shortest dependency paths and their closest neighbours in the learned feature space.}\label{tab:4}
\end{table}

From Table \ref{tab:4}, we can see that our approach can capture the core meaning of the shortest dependency paths.
For example, for the \textit{Instrument-Agency} relation, we infer that the dependency relations ``\textit{nsubj\_inv}'', ``\textit{dobj}'' and ``\textit{prep\_with}'' in the dependency path play a main role in the representation and our model can capture these similar paths.
For the \textit{Product-Producer} relation, our model focuses on representing the structure of ``\textit{nsubj\_inv} verb1 \textit{xcomp} verb2 \textit{dobj}'' and exploits some words like ``pencil'' and ``create'' in the path representation.
This is clearer for the \textit{Message-Topic} relation, where the similarity of words like ``point'', ``explore'', ``address'' and ``relate'' are well learned.

\subsubsection{Influence of Attached Subtree}
In this subsection, we will discuss the role of attached subtree (SUB) in relation classification.
 By comparing the results of DepNN before and after adding the subtree, we find the influence of this structure varies from different relation types. Table \ref{tab:SUB} shows the  F1 measures of each relation type before and after adding the subtree.
 \begin{table}[!htbp]
	\center
	\small

	\begin{tabular}{|l|ccc|}
		\hline
		\multirow{2}{*}{Relation}&
		\multicolumn{3}{c|}{F1}\\
		\cline{2-4}
		& No SUB & With SUB &Change\\
		\hline
Component-Whole&0.805&0.812&0.007\\
Instrument-Agency&0.683&0.714&0.031\\
Member-Collection&0.818&0.829&0.011\\
Cause-Effect&0.881&0.89&0.009\\
Entity-Destination&0.862&0.869&0.007\\
Content-Container&0.826&0.828&0.002\\
Message-Topic&0.854&0.856&0.002\\
Product-Producer&0.776&0.801&0.025\\
Entity-Origin&0.853&0.857&0.004\\
		\hline
	\end{tabular}
	\caption{Influence of the subtrees on each relation type.}\label{tab:SUB}
\end{table}

\begin{table*}
\center
\begin{tabular}{|l|l|cc|}
 \hline
 \multirow{2}{*}{Model} &
  \multirow{2}{*}{Additional Features (AF)} &
 \multicolumn{2}{c|}{F1}\\
  \cline{3-4}
   && with AF & without AF\\
 \hline
\multirow{4}{*}{SVM}&POS, prefixes, PropBank, Google n-gram,&\multirow{4}{*}{82.2}&\multirow{4}{*}{-}\\
&NomLex-Plus, Levin classes, WordNet,&&\\
&dependency parse, morphological,&&\\
&FrameNet, TextRunner, paraphrases&&\\
\hline
MV-RNN&POS, NER, WordNet&81.8\footnotemark[2]&78.2\\
\hline
CNN&WordNet&82.7&79.2\\
\hline
FCM&NER&83.0&82.2\\
\hline
DT-RNN&NER&73.1&72.1\\
\hline
DepNN&NER&\textbf{83.6}&\textbf{82.8}\\
\hline
 \end{tabular}
 \caption{Results of evaluation on the SemEval-2010 dataset.}\label{tab:results}
\end{table*}

\begin{figure}[!htbp]
\centering
\subfloat[]{
\label{fig:4_subfig_a}
\centering
\includegraphics[width=2.5in]{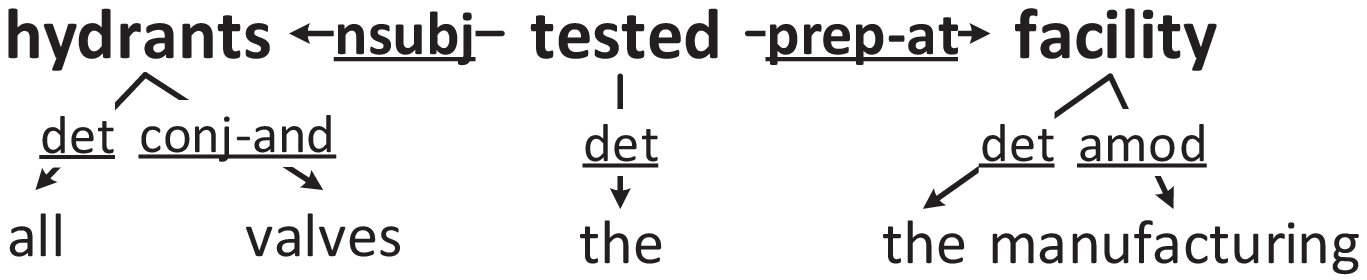}
}\\
\centering
\subfloat[]{
\label{fig:4_subfig_b}
\centering
\includegraphics[width=2.8in]{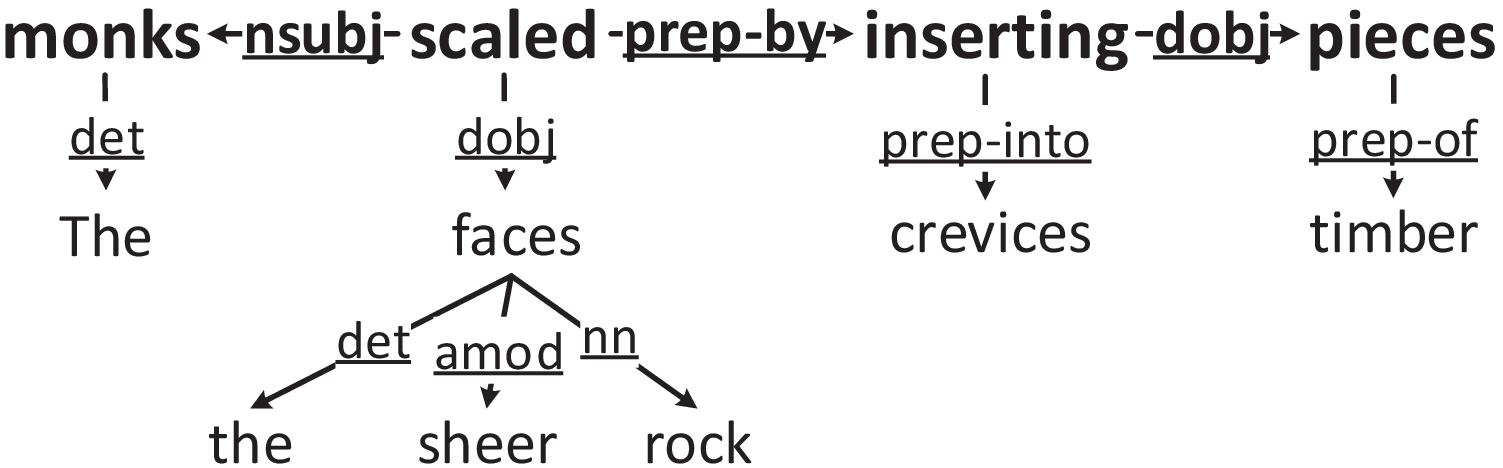}
}
\caption{ADP of instances that can be classified correctly after adding the subtrees.}
\label{fig:4_fig}
\end{figure}
 We can see that the subtree information generally has a positive impact on all the relation types. It is especially salient for the \textit{Instrument-Agency} and  \textit{Product-Producer} relations.
 With only using the shortest dependency paths, these two kinds of relation types are easily confused, as they both rely on the dependency paths such as ``$\dots$ verb \textit{prep-by/prep-with/using $\dots$}''.
But after considering the subtree information, we can better distinguish these two relation types. Figure \ref{fig:4_fig} lists two  instances that can be classified correctly only after adding the subtrees.  Figure \ref{fig:4_subfig_a} belongs to the \textit{Producer-Produce} relation which can be reflected by the subtree structures  like ``conj-and$\rightarrow$valves'' and ``amod$\rightarrow$manufacturing''.
Figure \ref{fig:4_subfig_b} belongs to the \textit{Instrument-Agency} relation, and the subtree structure attached to the word ``scaled'' provides more supplemental information to the shortest path as explained above.

\footnotetext[2]{MV-RNN achieves a higher F1-score (82.7) on SENNA embeddings reported in the original paper.}

\subsection{Comparison with Baselines}
In this subsection, we compare DepNN with several baseline approaches of relation classification.

{\bf SVM} \cite{UTD2010SemEval}:
This is the top performed system in SemEval-2010. It depends on the human compiled feature templates and then utilizes
many external corpora to extract features for an SVM classifier.

{\bf MV-RNN} \cite{socher2012emnlp}:
This model associates each word with a matrix. Based on the constituent parse tree structure, this model finds the path between two entities and learns the distributed representation of their highest parent node through the composition in a recursive neural network.

{\bf DT-RNN}~\cite{socher2014grounded} :
This model uses an RNN for modeling dependency trees. It assigns a composition matrix to each dependency relation.
Different from our model \textbf{DepNN}, the embedding of each node is a linear combination of its children.
The network is trained using the method provided by~\cite{iyyer2014neural}. We average the learned vectors of all nodes, stack it with the root node's embedding  and additional features, and feed them into a $softmax$ classifier.

{\bf CNN}:
\newcite{kang2014coling} build a convolutional model to learn a sentence representation over the words in a sentence.
To represent each word, they use a special position vector to indicate the relative distances of current input word to two marked entities,
concatenating the position vector with the corresponding word embedding.
Then the sentence representation  is  staked with some lexical features and fed into a $softmax$ classifier.

{\bf FCM} \cite{MoYu2014qv}:
FCM decomposes a sentence into some substructures and learns substructure embedding from each of them.
Then the substructure embeddings in a sentence are combined via a sum-pooling operation and put into a $softmax$ classifier.

Table~\ref{tab:results} compares DepNN with the baseline approaches.
Since many of our baselines are neural network models, it is convenient for them to use some features extracted with external resources or tools to enhance performance. We call these features ``additional features'' (AF) and list them in the second column.  The F1-measures on SemEval-2010 dataset  with/out these additional features are shown in the last two columns.

From Table~\ref{tab:results}, we can see that DepNN achieves the best result (83.6) with the NER  features.
SVM achieves a comparable result, though the quality of feature engineering highly relies on human experience and external NLP resources.
MV-RNN  models the constituent parse trees with a recursive procedure and its F1-measures with/out AF are about 1.7 percent and 4.6 percent lower than those of DepNN.
This to some extent indicates that our proposed  ADP structure is  more suitable for relation classification task.
Meanwhile, MV-RNN is very slow to train, since each word is associated with a matrix.
Both CNN and FCM use features from the whole sentence and achieve similar performance.
DT-RNN is the worst of all  baselines, though it also considers the information from shortest dependency paths and attached subtrees.
As we analyze, shortest dependency paths and subtrees play different roles in relation classification.
But, we can see that DT-RNN does not distinguish the modeling processes of shortest paths and subtrees, and deems the representation of each node as a linear combination of its children.
%This phenomenon is also seen in a kernel-based method~\cite{wang2008re}, where the tree kernel performs worse than the shortest path kernel.

 \setcounter{footnote}{2}
\section{Related Work}
Relation classification is  one traditional subproblem of Information Extraction (IE). It aims to detect and classify relations between the predefined types of objects in the corpus. These objects could be named entities or marked nominals\footnote{ACE Evaluation uses the named entities while the SemEval evaluation is based on nominals.}. Much research has been performed in this field, most of which considers it as a supervised multi-classification task. Depending on the input to the classifier, these approaches can be further divided into feature-based, tree kernel-based and composite kernel-based.

Feature-based methods extract various kinds of linguistic features, including both syntactic features and semantic cues. These features are combined to form a feature vector employed in a Max Entropy \cite{5219721} or an SVM \cite{2415182,guodong2005exploring} classifier. Feature-based methods usually need hand-crafted features and lack the ability to represent structural information (e.g., parsing tree, word order).

Kernel methods use a more natural way of exploring structural features by computing the inner product of two objects in the high-dimensional latent feature space. \newcite{794571} designed a tree kernel to compute the structural commonality between shallow parse trees by a weighted sum of the number of common subtrees. \newcite{1766597} transferred this kernel to a dependency tree and attached more information including POS tag, word chunk tag to each node. \newcite{zhou122007tree} proposed a context-sensitive convolution tree kernel that used context information beyond the local tree.
In another view, \newcite{2363908} provided an important insight that the shortest path between the two entities concentrates most of the information for identifying the relation between them. \newcite{nguyen2007relation} used the dependency subtrees in a different manner by modeling the subtrees between entities and keywords of certain relations.
\newcite{2415179} further proposed composite kernels to combine a tree kernel and a feature-based kernel to promote the performance.

Recently,  Deep Neural Networks (DNN) have been developed to solve the relation classification problem.
By associating each word a distributed representation, DNN can overcome the sparsity problem in traditional methods and automatically learn appropriate features. \newcite{socher2012emnlp} proposed a recursive neural network model by constructing compositional semantics for the minimal constituent of a constituent parse tree including both marked entities.
\newcite{kang2014coling} used a convolutional neural network over the whole sentence combined with some lexical features. They also pointed out that the position of each word in the sentence is very important for relation classification and concatenated a special position feature vector with the corresponding word embedding.
\newcite{MoYu2014qv} proposed the Factor-based Compositional Embedding Model which extracted features from the substructures of a sentence and combined them through a sum-pooling layer.

\section{Conclusion}
In this paper, we propose to classify relations between entities by modeling the augmented dependency path in a neural network framework.
For a given instance, we generate its ADP by combining the shortest path between two entities and the attached subtrees.
We present a novel approach, DepNN, to taking advantages of both convolutional neural network and recursive neural network to model this structure. Experiment results demonstrate that DepNN achieves state-of-the-art performance.

\bibliographystyle{acl}
\bibliography{paper}

\begin{thebibliography}{}

\bibitem[\protect\citename{Bach and Badaskar}2007]{bach2007survey}
Nguyen Bach and Sameer Badaskar.
\newblock 2007.
\newblock A survey on relation extraction.
\newblock {\em Language Technologies Institute, Carnegie Mellon University}.

\bibitem[\protect\citename{Bottou}2004]{bottou2004stochastic}
L{\'e}on Bottou.
\newblock 2004.
\newblock Stochastic learning.
\newblock In {\em Advanced lectures on machine learning}, pages 146--168.
  Springer.

\bibitem[\protect\citename{Bunescu and Mooney}2005]{2363908}
Razvan~C. Bunescu and Raymond~J. Mooney.
\newblock 2005.
\newblock {A Shortest Path Dependency Kernel for Relation Extraction}.
\newblock In {\em North American Chapter of the Association for Computational
  Linguistics}.

\bibitem[\protect\citename{Collobert \bgroup et al.\egroup
  }2011]{collobert2011natural}
Ronan Collobert, Jason Weston, L{\'e}on Bottou, Michael Karlen, Koray
  Kavukcuoglu, and Pavel Kuksa.
\newblock 2011.
\newblock Natural language processing (almost) from scratch.
\newblock {\em The Journal of Machine Learning Research}, 12:2493--2537.

\bibitem[\protect\citename{Culotta and Sorensen}2004]{1766597}
Aron Culotta and Jeffrey~S. Sorensen.
\newblock 2004.
\newblock {Dependency Tree Kernels for Relation Extraction}.
\newblock In {\em Meeting of the Association for Computational Linguistics},
  pages 423--429.

\bibitem[\protect\citename{de Marneffe and Manning}2008]{4933584}
Marie-Catherine de~Marneffe and Christopher~D. Manning.
\newblock 2008.
\newblock {The Stanford typed dependencies representation}.
\newblock In {\em International Conference on Computational Linguistics}.

\bibitem[\protect\citename{GuoDong \bgroup et al.\egroup
  }2005]{guodong2005exploring}
Zhou GuoDong, Su~Jian, Zhang Jie, and Zhang Min.
\newblock 2005.
\newblock Exploring various knowledge in relation extraction.
\newblock In {\em Proceedings of the 43rd annual meeting on association for
  computational linguistics}, pages 427--434. Association for Computational
  Linguistics.

\bibitem[\protect\citename{Hendrickx \bgroup et al.\egroup }2010]{5009814}
Iris Hendrickx, Zornitsa Kozareva, Preslav Nakov, Sebastian Pad~´ ok, Marco
  Pennacchiotti, Lorenza Romano, and Stan Szpakowicz.
\newblock 2010.
\newblock {SemEval-2010 Task 8: Multi-Way Classification of Semantic Relations
  Between Pairs of Nominals}.

\bibitem[\protect\citename{Iyyer \bgroup et al.\egroup }2014]{iyyer2014neural}
Mohit Iyyer, Jordan Boyd-Graber, Leonardo Claudino, Richard Socher, and
  Hal~Daum{\'e} III.
\newblock 2014.
\newblock A neural network for factoid question answering over paragraphs.
\newblock In {\em Proceedings of the 2014 Conference on Empirical Methods in
  Natural Language Processing (EMNLP)}, pages 633--644.

\bibitem[\protect\citename{Kambhatla}]{5219721}
Nanda Kambhatla.
\newblock {Combining Lexical, Syntactic, and Semantic Features with Maximum
  Entropy Models for Extracting Relations}.
\newblock In {\em Meeting of the Association for Computational Linguistics}.

\bibitem[\protect\citename{Klein and Manning}2003]{443415}
Dan Klein and Christopher~D. Manning.
\newblock 2003.
\newblock {Accurate Unlexicalized Parsing}.
\newblock In {\em Meeting of the Association for Computational Linguistics},
  pages 423--430.

\bibitem[\protect\citename{Nguyen \bgroup et al.\egroup
  }2007]{nguyen2007relation}
Dat~PT Nguyen, Yutaka Matsuo, and Mitsuru Ishizuka.
\newblock 2007.
\newblock Relation extraction from wikipedia using subtree mining.
\newblock In {\em Proceedings of the National Conference on Artificial
  Intelligence}, volume~22, page 1414. Menlo Park, CA; Cambridge, MA; London;
  AAAI Press; MIT Press; 1999.

\bibitem[\protect\citename{Rink and Harabagiu}2010]{UTD2010SemEval}
Bryan Rink and Sanda Harabagiu.
\newblock 2010.
\newblock Utd: Classifying semantic relations by combining lexical and semantic
  resources.
\newblock In {\em Proceedings of the 5th International Workshop on Semantic
  Evaluation}, pages 256--259, Uppsala, Sweden, July. Association for
  Computational Linguistics.

\bibitem[\protect\citename{Rumelhart \bgroup et al.\egroup
  }1988]{rumelhart1988learning}
David~E Rumelhart, Geoffrey~E Hinton, and Ronald~J Williams.
\newblock 1988.
\newblock Learning representations by back-propagating errors.
\newblock {\em Cognitive modeling}, 5.

\bibitem[\protect\citename{Socher \bgroup et al.\egroup }2012]{socher2012emnlp}
Richard Socher, Brody Huval, Christopher~D. Manning, and Andrew~Y. Ng.
\newblock 2012.
\newblock Semantic compositionality through recursive matrix-vector spaces.
\newblock In {\em Proceedings of the 2012 Joint Conference on Empirical Methods
  in Natural Language Processing and Computational Natural Language Learning},
  pages 1201--1211, Jeju Island, Korea, July. Association for Computational
  Linguistics.

\bibitem[\protect\citename{Socher \bgroup et al.\egroup
  }2014]{socher2014grounded}
Richard Socher, Andrej Karpathy, Quoc~V Le, Christopher~D Manning, and Andrew~Y
  Ng.
\newblock 2014.
\newblock Grounded compositional semantics for finding and describing images
  with sentences.
\newblock {\em Transactions of the Association for Computational Linguistics},
  2:207--218.

\bibitem[\protect\citename{Suchanek \bgroup et al.\egroup
  }2006]{suchanek2006combining}
Fabian~M Suchanek, Georgiana Ifrim, and Gerhard Weikum.
\newblock 2006.
\newblock Combining linguistic and statistical analysis to extract relations
  from web documents.
\newblock In {\em Proceedings of the 12th ACM SIGKDD international conference
  on Knowledge discovery and data mining}, pages 712--717. ACM.

\bibitem[\protect\citename{Wang}2008]{wang2008re}
Mengqiu Wang.
\newblock 2008.
\newblock A re-examination of dependency path kernels for relation extraction.
\newblock In {\em IJCNLP}, pages 841--846.

\bibitem[\protect\citename{Yu \bgroup et al.\egroup }2014]{MoYu2014qv}
Mo~Yu, Matthew Gormley, and Mark Dredze.
\newblock 2014.
\newblock Factor-based compositional embedding models.
\newblock In {\em NIPS Workshop on Learning Semantics}.

\bibitem[\protect\citename{Zelenko \bgroup et al.\egroup }2003]{794571}
Dmitry Zelenko, Chinatsu Aone, and Anthony Richardella.
\newblock 2003.
\newblock {Kernel Methods for Relation Extraction}.
\newblock {\em Journal of Machine Learning Research}, 3:1083--1106.

\bibitem[\protect\citename{Zeng \bgroup et al.\egroup }2014]{kang2014coling}
Daojian Zeng, Kang Liu, Siwei Lai, Guangyou Zhou, and Jun Zhao.
\newblock 2014.
\newblock Relation classification via convolutional deep neural network.
\newblock In {\em Proceedings of COLING 2014, the 25th International Conference
  on Computational Linguistics: Technical Papers}, pages 2335--2344, Dublin,
  Ireland, August. Dublin City University and Association for Computational
  Linguistics.

\bibitem[\protect\citename{Zhang \bgroup et al.\egroup }2006]{2415179}
Min Zhang, Jie Zhang, Jian Su, and Guodong Zhou.
\newblock 2006.
\newblock {A Composite Kernel to Extract Relations between Entities with Both
  Flat and Structured Features}.
\newblock In {\em Meeting of the Association for Computational Linguistics}.

\bibitem[\protect\citename{Zhou \bgroup et al.\egroup }2005]{2415182}
Guodong Zhou, Jian Su, Jie Zhang, and Min Zhang.
\newblock 2005.
\newblock {Exploring Various Knowledge in Relation Extraction}.
\newblock In {\em Meeting of the Association for Computational Linguistics}.

\bibitem[\protect\citename{Zhou \bgroup et al.\egroup }2007]{zhou122007tree}
GuoDong Zhou, Min Zhang, Dong~Hong Ji, and Qiaoming Zhu.
\newblock 2007.
\newblock Tree kernel-based relation extraction with context-sensitive
  structured parse tree information.
\newblock {\em EMNLP-CoNLL 2007}, page 728.

\end{thebibliography}

\end{document}